\documentclass[10pt,twocolumn,letterpaper]{article}

\usepackage{wacv}

\usepackage{times}
\usepackage{epsfig}
\usepackage[table]{xcolor}
\usepackage{graphicx}
\usepackage{amsmath}
\usepackage{amssymb}
\usepackage{dblfloatfix} 
\usepackage[pagebackref=true,breaklinks=true,letterpaper=true,colorlinks,bookmarks=false]{hyperref}
\usepackage{multirow}

\def\wacvPaperID{****} % Enter the WACV Paper ID here

\wacvfinalcopy % *** Uncomment this line for the final submission

\ifwacvfinal
\def\assignedStartPage{1} % *** Enter the assigned starting page number (instead of 9876)
\fi

\ifwacvfinal
\usepackage[breaklinks=true,bookmarks=false]{hyperref}
\else
\usepackage[pagebackref=true,breaklinks=true,colorlinks,bookmarks=false]{hyperref}
\fi

\ifwacvfinal
\else
\fi

\begin{document}

%%%%%%%%% TITLE

\title{FaceQvec: Vector Quality Assessment for Face Biometrics\\ based on ISO Compliance}

\author{Javier Hernandez-Ortega, Julian Fierrez, Luis F. Gomez and Aythami Morales\\
School of Engineering, Universidad Autonoma de Madrid, Spain\\
{\tt\small javier.hernandezo@uam.es, julian.fierrez@uam.es, luisf.gomez@uam.es, aythami.morales@uam.es}

\and
Jose Luis Gonzalez-de-Suso and Francisco Zamora-Martinez\\
Veridas Digital Authentication Solutions, Pamplona, Spain\\
{\tt\small jlgonzalez@veridas.com, pzamora@veridas.com}
}

\maketitle
%\thispagestyle{empty}

%%%%%%%%% ABSTRACT
\begin{abstract}
   In this paper we develop FaceQvec, a software component for estimating the conformity of facial images with each of the points contemplated in the ISO/IEC 19794-5, a quality standard that defines general quality guidelines for face images that would make them acceptable or unacceptable for use in official documents such as passports or ID cards. This type of tool for quality assessment can help to improve the accuracy of face recognition, as well as to identify which factors are affecting the quality of a given face image and to take actions to eliminate or reduce those factors, e.g., with postprocessing techniques or re-acquisition of the image. FaceQvec consists of the automation of $25$ individual tests related to different points contemplated in the aforementioned standard, as well as other characteristics of the images that have been considered to be related to facial quality. We first include the results of the quality tests evaluated on a development dataset captured under realistic conditions. We used those results to adjust the decision threshold of each test. Then we checked again their accuracy on a evaluation database that contains new face images not seen during development. The evaluation results demonstrate the accuracy of the individual tests for checking compliance with ISO/IEC 19794-5. FaceQvec is available online\footnote{https://github.com/uam-biometrics/FaceQvec}.
\end{abstract}

%%%%%%%%% BODY TEXT
\section{Introduction}

The accuracy of the output decision of a biometric system, e.g., a face recognition system, can only be as high as the reliability of its input data. That reliability, a concept popularly known as ``quality'', refers to the ability of the input sample to be used for recognition purposes producing accurate results~\cite{2011_QualityBio_FAlonso}. Recently, with the growth of biometrics, quality assessment has become one of the research topics with the highest interest of the community as it is one of the main factors responsible for the good performance of biometric systems~\cite{schlett2021face}. 

Simplifying the theory under biometric quality~\cite{grother2007performance}, we can state that if the input samples of a given biometric system are of low quality, the output that it will return is going to be inexact. On the other hand, if the input samples are of high quality, the results obtained will be more accurate. At this point one of the main drawbacks of face recognition appears, i.e., the high variability of the samples of a same subject due to the variability inherent to the acquisition process (done at a distance, under uncontrolled illumination, etc.). Due to that variability, the quality of face images can be really diverse, compromising the accuracy of face-based recognition~\cite{NIST_FRVT_QUALITY_2021}. 

%Examples of face images with diverse quality levels can be seen in Figure~\ref{fig:quality_examples}.

Knowing the impact that face quality has in the recognition accuracy, a big question arises: How can face quality be measured? Or more specifically: How can we distinguish between high quality and low quality images? In this work we focused on developing and evaluating FaceQvec, a bank of tests which return scores that serve as estimations of face quality. These multiple scores can be useful to know if a face image will give accurate results when used for face recognition. The selection of the tests that conform FaceQvec has been based in previous research in face and image quality and it contains methods to evaluate factors like: pose, illumination, blur, and occlusions.

%\begin{figure}[t]
%\centering
%\includegraphics[width=0.8\columnwidth]{variability_examples.eps} 
%\caption{\textbf{Examples of face images of different quality} from the VGGFace2 database. The top left image shows a gallery high quality image of a random subject while the other images are examples of low quality images from the same subject that suffer from diverse variability factors.}
%\label{fig:quality_examples}
%\end{figure}

In this paper we: i) present FaceQvec, a new face quality assessment software based on a collection of $25$ individual tests designed to check image compliance with the ISO/IEC 19794-5 standard, ii) include a brief description of each one of the $25$ tests that compose FaceQvec, iii) test FaceQvec on a dataset acquired under realistic conditions in order to adjust the decision thresholds of the different tests, and iv) we evaluate the accuracy of the adjusted tests on another realistic dataset. Our results validate the utility of FaceQvec, and open new application opportunities to face quality assessment methods. This proposed approach is in line with the recent announcement by NIST of a new benchmark campaign focused in vector quality assessment for face biometrics related to ISO compliance. This new NIST initiative follows another related very successful benchmark campaign around scalar face biometric quality that have been running in the last 2 years~\cite{NIST_FRVT_QUALITY,NIST_FRVT_QUALITY_2021}, and it is related to the standard on face biometric quality ISO/IEC WD 29794-5 now under development.

The rest of this paper is organized as follows: Section~\ref{sec:related_works} gives an overview of the field of face quality and the works proposed so far in the literature. Section~\ref{sec:development} describes FaceQvec, including the description of the tests that conform the software. Section~\ref{sec:evaluation} summarizes the development and evaluation databases, the experimental protocol, and the results obtained. Finally, concluding remarks and future work are drawn in Section \ref{sec:conclusion}.

\section{Related works}
\label{sec:related_works}

Nowadays, two of the most relevant and extended public standards related to quality assessment in biometrics are the ICAO 9303 and the ISO/IEC 19794-5~\cite{REFstandard2011}. These documents are actually a series of guidelines for the acquisition of high quality images, i.e., portrait-like images, for their inclusion in machine-readable official documents like passports and ID cards. These guidelines are based on the typical impact that certain features like blur, occlusions, and resolution have in the quality of facial images. However, these reports do not specify the minimum requirements for each of the quality features in order to consider an image of high quality, and they neither indicate the method to measure each of the features. In order to implement their recommended guidelines for face quality assessment, it becomes necessary to define specific tests and minimum thresholds for each one of the quality features that can be used to verify the compliance with the standards. 

%In the last decade some vendors and academic works have developed automatic tools to check if a face image complies or not with the quality guidelines. These tools usually output a binary vector in which each position defines whether or not a specific test is passed/not-passed by the image.

%The ICAO 9303 and the ISO/IEC 19794-5 standards have not changed since its publication until today, so the problem related to their ambiguity is still present.

First works related to face image quality assessment appeared at the beginning of the 00's~\cite{weber2006some,gao2007standardization}, and were generally centered in extracting hand-crafted features from face images and using them to calculate one or a few quality measures~\cite{beveridge2008focus,phillips2013existence,raghavendra2014automatic}. These measures estimate the presence of features like illumination, blurriness, and extreme pose, that can have a significant impact on the recognition performance. A good example of these approaches is the work in~\cite{beveridge2010quantifying}, where the authors focused on studying the impact that different levels of illumination can have on face recognition. 

%They concluded that the accuracy of some of the most used face recognition algorithms were really affected by that quality factor.

The main drawback behind these first approximations is their narrow scope as they only measured one or two quality features at most. Another handicap is that many of these earlier methods did not return a numerical value for each test that can be used to establish thresholds to decide if an image complies with the standard or not. A more recent work~\cite{abaza2012quality} took a step forward to solve those limitations by augmenting the number of features they measured and computing a numerical Face Quality Index (FQI) that combines five individual quality factors: contrast, brightness, focus, sharpness, and illumination. 

%Then they defined the final FQI normalising each quality measure and modeling their distributions as Gaussian PDFs, where values close to the mean of each PDF correspond high quality and scores far to the mean correspond to low quality.

The BioLab-ICAO framework was presented in~\cite{ferrara2012face} as an evaluation tool for ISO/IEC compliance checking. The authors, after an in-depth study of the ISO/IEC 19794-5 standard, defined a set of $30$ different individual tests for each input image related to the geometry of the face, e.g., location and separation of the eyes, and to the photographic properties of the images, e.g., focus and contrast. The output of each one of those tests consists of a numerical score in the [$0$,$100$] range. This framework represented one of the first attempts of developing an automatic tool to assess the level of compliance of an image with a public standard in face biometrics.

The high growth experienced by deep learning in the last decade, mainly due to its improved accuracy respect to hand-crafted methods, has led the research linked to face quality assessment to also adopt these methods with great success. This is the case of works like ~\cite{zhang2017illumination,wang2017learning} where Convolutional Neural Networks (CNNs) where used to predict the presence of factors like the quality of the illumination.

Most of the current works in face quality assessment are following a different approach compared to the older works mentioned previously in this section. The early research stage was mainly focused on developing individual tests capable of giving an estimation of the presence of factors that researchers assume that should affect face quality, e.g., blur, resolution, and occlusions. Works like~\cite{best2018learning,hernandez2020biometric,2021_FaceQgen_Hernandez} have the objective of correlating the quality of an image to the expected accuracy when using that specific sample for face recognition. They do it by training deep learning models using large datasets labeled with quality values related to face recognition. Thus, the predictions from the trained models will be highly correlated to the face recognition accuracy of some state-of-the-art commercial recognition systems. These new approaches to face quality assessment are very useful to improve the performance of face recognition systems, since for each input image they return a global numerical quality score specially designed with that target in mind. However, the methods designed to that extent do not usually give information about which specific image features are affecting the quality of the face images. Knowing about those individual image features can be very beneficial, e.g., in the enrollment of new users, when detecting the presence of bad quality factors can be used to give detailed feedback to the users to solve the acquisition problems that may be occurring.

%, e.g., sensor malfunction, dirtiness in the sensor, bad user positioning, etc.

%However, in the last couple of years, researchers are more focused on producing quality measures related to recognition accuracy, that can be complex and not related to single image features. 

%Even though knowing the expected performance of face recognition is critical in real application scenarios, 

With this desire in mind, our target here has been developing FaceQvec, a software tool that comprises a selection of face quality tests as complete as possible. We took the work in~\cite{ferrara2012face} as our main reference since, as far as we know, there are no other works so comprehensive, with such a good scientific basis, and so well documented regarding the study of the compliance of facial images with the ISO/IEC standard. However, there is more advanced and accurate technology at present than the one used in~\cite{ferrara2012face} in 2012 that would allow automating some of the quality tests with much better accuracy. For example, recent advances in deep learning and computer vision could be used to obtain more robust results regarding tests like pose estimation and eye tracking, that in 2012 were carried out with slower and less reliable algorithms.

Summarizing, the present work develops an automated tool for face quality estimation based on the ISO/IEC 19794-5 standard. We developed a software component consisting of a set of $25$ individual tests based on the guidelines of the mentioned standard. We used a combination of traditional methods (known in the literature as hand-crafted) and deep learning-based models to calculate the result of each one of the tests. The quality measures from FaceQvec can be used to improve face recognition in several ways: \textit{i)} discarding the samples that do not reach a minimum quality during enrollment; \textit{ii)} estimating the reliability of face recognition when using a specific sample for that purpose; and \textit{iii)} as a confidence value to improve decision-making processes~\cite{new9}. 

\begin{table*}
\caption{\textbf{Definition of the quality tests of FaceQvec} including brief descriptions of the methods used.}
\centering
\resizebox{0.9\linewidth}{!}{
\begin{tabular}{|c|l|l|}
\hline
\cellcolor[HTML]{C0C0C0}\textbf{Test}   & \textbf{Description} & \textbf{Method}\\
\hline
\hline
\cellcolor[HTML]{C0C0C0}1 & Blur & Laplacian of the image to highlight the edges in it.\\
\hline
\cellcolor[HTML]{C0C0C0}2 & Eyes direction & Euclidean distance between the pupil and the center of the eyeball.\\
\hline
\cellcolor[HTML]{C0C0C0}3 & Presence of ink marks & Background and face segmentation and a color-based ink detector.\\
\hline
\cellcolor[HTML]{C0C0C0}4 & Odd skin colour & Detection of skin pixels of unnatural colour based on a color-based detector.\\
\hline
\cellcolor[HTML]{C0C0C0}5 & General illumination & Detecting if the facial image is too dark (or too bright) based on mean pixels value.\\
\hline
\cellcolor[HTML]{C0C0C0}6 & Contrast & Checking if the pixels are concentrated in a small part of the possible range.\\
\hline
\cellcolor[HTML]{C0C0C0}7 & Pixelation & Checking the presence of horizontal and vertical borders at a periodic distance.\\
\hline
\cellcolor[HTML]{C0C0C0}8 & Hair over face & Hair segmentation inside the face zone using a pretrained CNN.\\
\hline
\cellcolor[HTML]{C0C0C0}9 & Eyes open/closed & Measuring the distance between the landmarks of the eyes.\\
\hline
\cellcolor[HTML]{C0C0C0}10 & Heterogeneous background & Background segmentation and clustering of the pixels' values using k-means.\\
\hline
\cellcolor[HTML]{C0C0C0}11 & Pose estimation & Roll, pitch, and yaw estimation using a pretrained CNN.\\
\hline
\cellcolor[HTML]{C0C0C0}12 & Light reflections on skin & Looking for overexposed zones inside the face based on pixels' values.\\
\hline
\cellcolor[HTML]{C0C0C0}13 & Red eyes & Looking for red eyes based on colour segmentation.\\
\hline
\cellcolor[HTML]{C0C0C0}14 & Shadows in the background & Background segmentation and shadow detection based on colour.\\
\hline
\cellcolor[HTML]{C0C0C0}15 & Shadows over face & Face segmentation and shadow detection based on colour.\\
\hline
\cellcolor[HTML]{C0C0C0}16 & Detection of sunglasses & Detection of dark pixels in the eyes region and its surroundings.\\
\hline
\cellcolor[HTML]{C0C0C0}17 & Light reflections on glasses & Looking for overexposed pixels in the eyes region and its surroundings.\\
\hline
\cellcolor[HTML]{C0C0C0}18 & Wide frames of the glasses & Looking for wide edges in the eye zone surroundings.\\
\hline
\cellcolor[HTML]{C0C0C0}19 & Frames covering the eyes & Looking for edges inside the eye zone.\\
\hline
\cellcolor[HTML]{C0C0C0}20 & Hat & Looking for pixels if unnatural colour in the upper forehead region.\\
\hline
\cellcolor[HTML]{C0C0C0}21 & Veil & Looking for pixels if unnatural colour in the lower part of the face.\\
\hline
\cellcolor[HTML]{C0C0C0}22 & Mouth open/closed & Measuring the distance between the landmarks of the mouth.\\
\hline
\cellcolor[HTML]{C0C0C0}23 & Other faces & Detecting if there are other faces in the images.\\
\hline
\cellcolor[HTML]{C0C0C0}24 & White noise estimation & Convolution with a kernel designed to remark this type of noise.\\
\hline
\cellcolor[HTML]{C0C0C0}25 & Expression & Detecting subject's facial expression using a pretrained CNN.\\
\hline
\end{tabular}
\label{table:tests}
}
\end{table*}

\section{FaceQvec: framework description}
\label{sec:development}

This section enumerates and describes the set of individual quality tests that conform FaceQvec, which have been designed to evaluate the compliance of a face image with the ISO/IEC 19794-5 standard that indicates some of the factors that can affect face quality. According to it, controlling elements such as resolution, illumination, pose, and focus will make two images coming from the same subject to look as similar as possible. These are the kind of images that can be consider of high quality since they will make easier to distinguish the identity of the person in the photo, therefore increasing the accuracy of face recognition. However, as we mentioned previously, the standard does not contain descriptions of all the quality factors that can affect a face image, and for those that it describes it does not specify a concrete way to measure them.

Due to the lack of details of the ISO/IEC standard and its ambiguity, the first step we considered when designing FaceQvec was defining the selection of the individual features that we want to measure for each face image. This selection had to be as complete as possible in order to serve as a reliable estimation of face image quality. Initially, after a through review of the literature (that we have summarized in Section~\ref{sec:related_works}), we decided to use the the work presented in~\cite{ferrara2012face} as the main reference of our work in order to make the initial selection of the individual quality tests to be included in FaceQvec. We decided to do so because the tests described in~\cite{ferrara2012face} are of the same nature and objectives as those of the present project. 

The authors of~\cite{ferrara2012face}, after studying the ISO/IEC standard, defined a set of $30$ well-defined features related to the geometry of the face, e.g., location and separation of the eyes, and to the photographic properties of the images, e.g., focus and contrast. Those characteristics would serve as criteria to assess the degree of conformity of a facial image with the ISO/IEC quality standard. In this project we have started from the proposal in \cite{ferrara2012face}, but doing a different selection of tests and applying more accurate technologies and algorithms when possible.

\subsection{Selection of the final set of quality tests}

The criteria we followed for selecting the final set of quality features to be measured has been based on the following points: 1) the expected impact that each feature can have on face recognition (based on the literature), 2) the computational requirements of measuring each quality feature, and 3) the existence of well-known algorithms/methods to calculate each feature, or alternatively, the easiness of implementing our own solution.

Based on the enumerated criteria, from all tests the defined in~\cite{ferrara2012face} we made a first selection of candidates, focusing on those classified in their paper as ``Photographic and pose-specific tests'', discarding a few of them, and also those that use to be carried out by other modules/stages of the recognition pipeline like the face detector. Then this first selection has been completed with tests from other relevant publications in the literature and also with additional self-designed tests for checking image features that, based on our experience, can severely affect the quality of facial images. After this definition stage, we concluded that a total of $25$ tests would be sufficient to evaluate the compliance of a facial image with the ISO/IEC standard.

\subsection{FaceQvec implementation details}
\label{sec:implementation}

The first stage in the workflow of FaceQvec consists in a preprocessing phase that is applied to input images in order to normalize and regularize them. Some of the tests are highly sensitive to changes in the images like their size or the precision of the face detector, so this stage was added to make their results as robust as possible. The three steps that conform the preprocessing module are:

\begin{itemize}
    \item \textbf{Face detection and localisation}: we use a face detection engine based on MobileNet v2, fine-tuned for face detection using a private database of our own. \\
    
    The outputs of the detector are the coordinates of the bounding boxes encompassing each of the faces detected in the image.

    \item \textbf{Face cropping}: we discard the image outside the bounding box but we add a margin of $20$ pixels on each side of the bounding box (when possible) as many detectors crop part of the ears, chin, and hair. Some of the quality tests are focused on studying these areas, so including them in the final image is necessary.

    \item \textbf{Face Resizing}: the final face image is resized to $112 \times 112 \times 3$ pixels. It is important to follow this last step, as many of the checks are sensitive to image size and resolution.
    
\end{itemize}

After the preprocessing stage, the resulting face images are evaluated on each one of the $25$ quality tests. A definition of each one of the tests, aside a brief description is given in Table~\ref{table:tests}. We want to highlight that for the implementation some of the tests, e.g., Mouth open, Eyes closed, Hair over face, and Pose estimation, we have applied deep learning models motivated by their higher accuracy compared to traditional methods in tasks such as image segmentation and landmarks detection.

The original output of each individual test was a number that could serve as an estimation of the level of compliance of the image with each specific point of the quality standard. However, this type of output represented difficulties for its evaluation since the methods' accuracy is computed on real databases where numerical and objective quality scores are usually not available. For example, the datasets we used in this paper are labeled by experts, not with numerical scores about the level of compliance but with binary decisions regarding ISO/IEC compliance for each considered quality feature. The images receive a $0$ if the experts consider they do not comply with a specific feature or they will be labeled with a $1$ if they have enough quality to comply with that part of the quality standard. Thus, to be able to compare the groundtruth quality labels with the outputs of the tests of FaceQvec we need the output of the tests to be also binary, i.e., for each specific facial image the tests must answer the question: Does the image pass the test? Consequently, we have thresholded the original output of each one of the individual tests to make them to return a binary score ($0$ or $1$) similar to the groundtruth labels. This way we manage to know categorically whether or not a given facial image meets the minimum quality requirements related to each aspect of the ISO/IEC standard.

With this approach we think we have managed to obtain an accurate estimation of the quality of face images both from the perspective of each individual quality feature and from a global point of view thanks to the complete selection of tests. It would be possible to extend these results in the future if we have access to a database with numerical labels that measure the degree of compliance with the test, e.g., with a numerical score between $0$ and $100$.

\section{FaceQvec: evaluation}
\label{sec:evaluation}

\begin{figure}[t!]
\centering
\includegraphics[width=0.8\columnwidth]{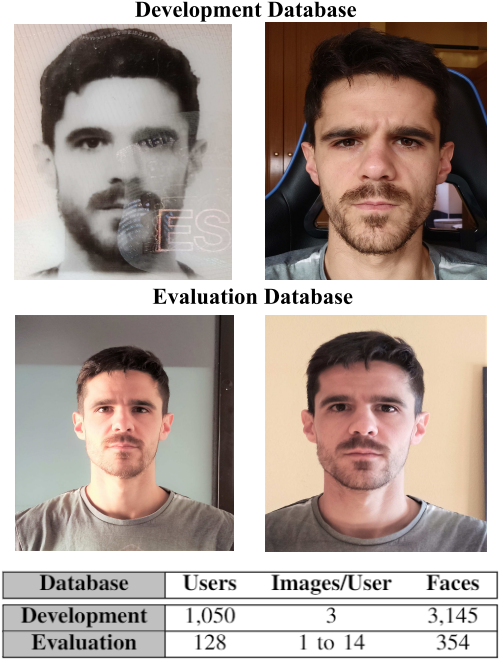} 
\caption{\textbf{Examples of images of the development and evaluation databases.} Upper row: ID and Selfie images of the development database. Bottom Row: images of the evaluation database.}
\label{fig:examples_of_databases}
\end{figure}

In this section we present two different evaluations made to the $25$ tests of FaceQvec. With these evaluations we intend to answer the question: How accurate is a particular quality test? For this purpose, a first analysis of the individual tests was done using a development database captured in a real scenario. With this evaluation, in addition to have a first glance of the accuracy we can expect from the tests, we were searching to adjust the configuration of their decision thresholds. Then, we performed a second evaluation with the optimized thresholds on a different database to see how discriminating the tests are when facing new and never-seen face images.

\subsection{Databases}

The development database is composed of $1$,$050$ subjects, containing $3$ face images for each one of them: $2$ front photographs of their official ID document (one taken with the flash activated and one without flash), and $1$ selfie photograph captured by the subject itself. All the images are labeled by experts with regard with their compliance with each quality test defined in FaceQvec (positive/negative decision). The database also includes the images after face detection and cropping. The images of this database were captured in a real scenario where the users of a mobile application were said to take photographs of their ID documents and also a selfie, and after that they had to upload the images to a server using the app. Therefore, the images of the database vary enormously in their quality as different users had cameras of distinct qualities, their skill as photographers was also variable, and the acquisition conditions (mostly of the selfie) could be more or less favourable, i.e., the lighting conditions, the presence of blur, etc. 

The second database consists of $320$ images of $128$ different subjects, with a ratio of images per subject ranging from $1$ to $14$ images. The images were captured in a scenario that mimics a border access control post. We built a vertical stand with an Intel RealSense DS435 camera on it. The images were captured under $4$ different illumination conditions: superior lighting, back lighting, frontal lighting, and diffuse lighting. Similarly to the development database, in this case the images are also labelled by humans according to their compliance with each one of the tests of FaceQvec. Nevertheless, the images of this database are of a different nature than the ones from the development database, presenting more controlled acquisition conditions so this should be translated into higher face quality values. The different illumination scenarios can also impact in the results of several quality tests like those related with shadows, overexposure, or skin color.

Figure~\ref{fig:examples_of_databases} shows the structure of the databases and also some examples of images belonging to the development and the evaluation databases.

\begin{table}[t]
\caption{\textbf{Distribution of groundtruth quality labels} for the development database.}
\centering
\resizebox{0.8\linewidth}{!}{
\begin{tabular}{|c|c|c|c|c|c|}
\hline
\cellcolor[HTML]{C0C0C0}\textbf{Test} & \textbf{Positive} & \textbf{Negative} & \cellcolor[HTML]{C0C0C0}\textbf{Test} & \textbf{Positive} & \textbf{Negative}\\
\hline
\hline
\cellcolor[HTML]{C0C0C0}\textbf{1} & 979 & 83 & \cellcolor[HTML]{C0C0C0}\textbf{14} & 1031 & \cellcolor[HTML]{FF5733}31\\
\hline
\cellcolor[HTML]{C0C0C0}\textbf{2} & 822 & 83 & \cellcolor[HTML]{C0C0C0}\textbf{15} & 632 & 430\\
\hline
\cellcolor[HTML]{C0C0C0}\textbf{3} & 1062 & \cellcolor[HTML]{FF5733}0 & \cellcolor[HTML]{C0C0C0}\textbf{16} & 1062 & \cellcolor[HTML]{FF5733}0\\
\hline
\cellcolor[HTML]{C0C0C0}\textbf{4} & 995 & 67 & \cellcolor[HTML]{C0C0C0}\textbf{17} & 969 & 93\\
\hline
\cellcolor[HTML]{C0C0C0}\textbf{5} & 887 & 175 & \cellcolor[HTML]{C0C0C0}\textbf{18} & 957 & 105\\
\hline
\cellcolor[HTML]{C0C0C0}\textbf{6} & 951 & 111 & \cellcolor[HTML]{C0C0C0}\textbf{19} & 1055 & \cellcolor[HTML]{FF5733}7\\
\hline
\cellcolor[HTML]{C0C0C0}\textbf{7} & 793 & 269 & \cellcolor[HTML]{C0C0C0}\textbf{20} & 1056 & \cellcolor[HTML]{FF5733}6\\
\hline
\cellcolor[HTML]{C0C0C0}\textbf{8} & 1037 & \cellcolor[HTML]{FF5733}25 & \cellcolor[HTML]{C0C0C0}\textbf{21} & 1056 & \cellcolor[HTML]{FF5733}6\\
\hline
\cellcolor[HTML]{C0C0C0}\textbf{9} & 1020 & \cellcolor[HTML]{FF5733}24 & \cellcolor[HTML]{C0C0C0}\textbf{22} & 883 & 179\\
\hline
\cellcolor[HTML]{C0C0C0}\textbf{10} & 692 & 370 & \cellcolor[HTML]{C0C0C0}\textbf{23} & 1062 & \cellcolor[HTML]{FF5733}0\\
\hline
\cellcolor[HTML]{C0C0C0}\textbf{11} & 929 & 133 & \cellcolor[HTML]{C0C0C0}\textbf{24} & 1055 & \cellcolor[HTML]{FF5733}7\\
\hline
\cellcolor[HTML]{C0C0C0}\textbf{12} & 953 & 109 & \cellcolor[HTML]{C0C0C0}\textbf{25} & 677 & 385\\
\hline
\cellcolor[HTML]{C0C0C0}\textbf{13} & 1057 & \cellcolor[HTML]{FF5733}5 & \cellcolor[HTML]{C0C0C0}- & - & -\\
\hline
\end{tabular}
\label{table:training_db_labels}
}
\end{table}

\begin{table}[t!]
\caption{\textbf{Distribution of groundtruth quality labels} for the evaluation database.}
\centering
\resizebox{0.8\linewidth}{!}{
\begin{tabular}{|c|c|c|c|c|c|}
\hline
\cellcolor[HTML]{C0C0C0}\textbf{Test} & \textbf{Positive} & \textbf{Negative} & \cellcolor[HTML]{C0C0C0}\textbf{Test} & \textbf{Positive} & \textbf{Negative}\\
\hline
\hline
\cellcolor[HTML]{C0C0C0}\textbf{1} & 314 & 40 & \cellcolor[HTML]{C0C0C0}\textbf{14} & 351 & \cellcolor[HTML]{FF5733}3\\
\hline
\cellcolor[HTML]{C0C0C0}\textbf{2} & 315 & 39 & \cellcolor[HTML]{C0C0C0}\textbf{15} & 347 & \cellcolor[HTML]{FF5733}7\\
\hline
\cellcolor[HTML]{C0C0C0}\textbf{3} & 345 & \cellcolor[HTML]{FF5733}9 & \cellcolor[HTML]{C0C0C0}\textbf{16} & 354 & \cellcolor[HTML]{FF5733}0\\
\hline
\cellcolor[HTML]{C0C0C0}\textbf{4} & 352 & \cellcolor[HTML]{FF5733}2 & \cellcolor[HTML]{C0C0C0}\textbf{17} & 299 & 55\\
\hline
\cellcolor[HTML]{C0C0C0}\textbf{5} & 349 & \cellcolor[HTML]{FF5733}5 & \cellcolor[HTML]{C0C0C0}\textbf{18} & 338 & 16\\
\hline
\cellcolor[HTML]{C0C0C0}\textbf{6} & 354 & \cellcolor[HTML]{FF5733}0 & \cellcolor[HTML]{C0C0C0}\textbf{19} & 345 & \cellcolor[HTML]{FF5733}9\\
\hline
\cellcolor[HTML]{C0C0C0}\textbf{7} & 352 & \cellcolor[HTML]{FF5733}2 & \cellcolor[HTML]{C0C0C0}\textbf{20} & 353 & \cellcolor[HTML]{FF5733}1\\
\hline
\cellcolor[HTML]{C0C0C0}\textbf{8} & 351 & \cellcolor[HTML]{FF5733}3 & \cellcolor[HTML]{C0C0C0}\textbf{21} & 346 & \cellcolor[HTML]{FF5733}8\\
\hline
\cellcolor[HTML]{C0C0C0}\textbf{9} & 348 & \cellcolor[HTML]{FF5733}6 & \cellcolor[HTML]{C0C0C0}\textbf{22} & 317 & 37\\
\hline
\cellcolor[HTML]{C0C0C0}\textbf{10} & 332 & 22 & \cellcolor[HTML]{C0C0C0}\textbf{23} & 354 & \cellcolor[HTML]{FF5733}0\\
\hline
\cellcolor[HTML]{C0C0C0}\textbf{11} & 345 & \cellcolor[HTML]{FF5733}9 & \cellcolor[HTML]{C0C0C0}\textbf{24} & 354 & \cellcolor[HTML]{FF5733}0\\
\hline
\cellcolor[HTML]{C0C0C0}\textbf{12} & 305 & 49 & \cellcolor[HTML]{C0C0C0}\textbf{25} & 231 & 123\\
\hline
\cellcolor[HTML]{C0C0C0}\textbf{13} & 351 & \cellcolor[HTML]{FF5733}3 & \cellcolor[HTML]{C0C0C0}- & - & -\\
\hline
\end{tabular}
\label{table:test_db_labels}
}
\end{table}

The process of acquiring the databases let us clear that quality assessment tools like FaceQvec are necessary in order to avoid the inclusion of ``garbage'' in the datasets during enrollment. In the case of the development database, many users took their selfies while wearing sunglasses, hats, and other facial complements, with extreme poses and angles, or under bad illumination conditions. A software like FaceQvec can be used during acquisition to determine in which way an image is not complying with the quality standard, and then that feedback can be given to the user to solve the problems in the image before acquiring a new one. In the case of the second database, where the images were captured simulating a border access control post, the face quality information can be used in the same way to give feedback to the border guards. 

As we mentioned previously, the images of both databases have been labeled by experts with binary information about the presence or absence of each one of the individual quality features evaluated by the tests. A \textbf{positive} label ($1$) means that the image complies with the ISO/IEC standard for that specific quality test. On the contrary, an image with a \textbf{negative} label ($0$) means that ISO/IEC would not give its approval for that individual quality test. Since both databases were captured in realistic environments, for most of the tests the number of images with positive and with negative labels will be unbalanced. 

Table~\ref{table:training_db_labels} shows the distribution of positive and negative labels for each individual test for the images of the development database. As can be seen in the table, we only run FaceQvec's quality tests on the selfie images of the development database, not on the ID photographs. FaceQvec has been designed mainly thinking about giving feedback to users. However, in the case of the photographs taken to the ID documents, we are actually facing face images included into the documents so improving their quality is not possible. We could use FaceQvec to detect some problems associated to general image quality, i.e., reflexes, blur, low contrast, and unnatural color, but there is nothing we can do with the quality of the face itself.

\begin{figure*}t!]
\centering
\includegraphics[width=0.8\linewidth]{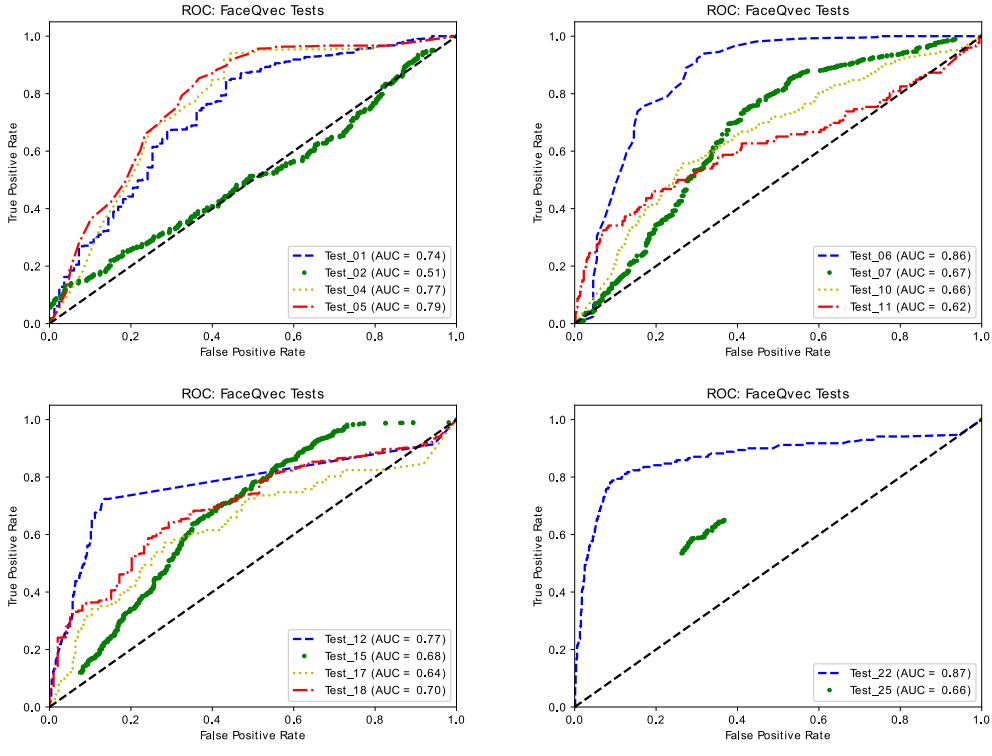} 
\caption{\textbf{ROC curves for the FaceQvec tests}. Results were obtained only for the selfie images of the development database and for those tests with a significant amount of negative cases.}
\label{fig:ROC_plots}
\end{figure*}

Table~\ref{table:training_db_labels} also shows that there are some quality tests with the two classes unbalanced, like the ones that measure the presence of red eyes, hats, and veils. All the tests with underrepresentation of negative samples in the development database are highlighted in red in the table. This underrepresentation for some tests was expected as the users were said to take their selfies trying to reach portrait-like quality. 

In the same manner, Table~\ref{table:test_db_labels} shows the number of images with positive and negative labels for the second database. In this case there are many more tests with unbalanced classes than for the development database. This is caused by the more controlled acquisition conditions and the lower number of images in the database. From the point of view of the results, this underrepresentation for some tests makes impossible to determine their error rates correctly. Due to this, results for certain quality tests are not included in this article and others are not completely reliable. In the future we plan to acquire and label a new database containing a more balanced number of samples for each of the quality tests proposed in FaceQvec.

\subsection{Analysis on development data}

In this section we analyse the results of a first evaluation of the quality tests over the development database (only for those tests with a significant number of both negative and positive samples). With this evaluation, in addition to obtaining a first estimation of the accuracy of the tests, we also wanted to use the information to adjust the individual decision thresholds to make the tests work with high accuracy when facing new images.

For each quality test of FaceQvec we calculated a Receiver Operating Characteristic curve (ROC) comparing the True Positive Rate (TPR), i.e., the percentage of images classified positively that are actually labeled positively in the development database, versus the False Positive Rate (FPR), i.e., the percentage of images estimated as positives that were actually labeled as negatives in the development database. This way we were able to fix a decision threshold for each test in a point with a good balance between sensitivity and tolerance. Each point of the ROC corresponds to a pair (TPR, FPR) obtained for a specific value of the decision threshold. For each curve we also computed the Area Under the Curve (AUC) as it is a useful metric to estimate the discriminating ability of the quality tests. An AUC closer to $1$ means that the test is highly accurate, since that implies that it has a TPR near to $1$ and a FPR near to $0$. We calculated these performance metrics for all the tests except for those highlighted in red in Table~\ref{table:training_db_labels}, due to the scarcity of negative samples. 

The ROCs together with the corresponding AUCs can be seen in Figure~\ref{fig:ROC_plots}. Most of the tests obtained a value for the AUC between $0.65$ and $0.80$, with a few ones like ``Mouth open'' that presents an even higher AUC ($0.87$ in this case). After looking at the results we made a classification of the tests into three different categories according to their AUC, i.e., their performance level: 1) \textit{High performance}: tests with AUC values equal or superior to $0.75$, 2) \textit{Medium performance}: those with an AUC between $0.65$ and $0.75$, and 3) \textit{Low performance}: tests with an AUC under $0.65$.

Due to extension constraints we can not discuss here the results for all the tests so decided to focus on the \textit{Low performance} class to try to reveal the causes of their high error rates: 

\begin{itemize}
    \item \textbf{Test 2 (Eyes direction)}: In this case the AUC was extremely low ($0.51$) being similar to a random decision. This test makes use of a pretrained CNN for the detection of the iris landmarks. It seems than that detection has been deficient and therefore, the distance between the geometric center of the eyeball and the pupil is not reliable enough. We think this can be caused by the low resolution of the images of the development database, since further testing made on images with higher resolution showed superior performance both for the landmark detection model and for the whole test.
    
    \item \textbf{Test 11 (Pose estimation)}: For this test, we think that the low accuracy (AUC = $0.62$) is also caused by a pretrained CNN, in this case used for pose estimation. The main issue with this pretrained model is related with the preprocessing stage of the input images that crops the images leaving only the face area. However, the pose estimation model needs information about the surrounding area of the face in order to make accurate estimations of the roll, pith, and yaw angles. Additional tests showed that the accuracy of this tests increases when executed before image cropping. 
    
    \item \textbf{Test 17 (Light reflections on glasses)}: This test has obtained a low AUC ($0.64$) likely due to the poor accuracy of the eye landmarks detection when the subjects are wearing glasses. Here we think that using an eye landmarks detector more robust to the presence of glasses will be helpful to increase the accuracy of the test.
    
\end{itemize}

After this analysis of the results, we used the ROC curves to fix the decision threshold of each individual quality test. We decided to set the thresholds to a value that gave us the highest TPR possible while maintaining the FPR at least under a $50$\%. These are the values of the thresholds that will be used later in the evaluation over the second dataset in order to see how well the quality tests generalise on new and different face images.

\subsection{Analysis on evaluation data}

\begin{table}
\caption{\textbf{Performance of the quality tests} for the evaluation database.}
\centering
\resizebox{0.6\linewidth}{!}{
\begin{tabular}{|c|ccc|}
\hline
\cellcolor[HTML]{C0C0C0}\textbf{Test} & \textbf{Accuracy}  & \textbf{TPR} & \textbf{FPR}\\
\hline
\hline
\cellcolor[HTML]{C0C0C0}\textbf{1} & 0.79 & 0.95 & 0.05 \\
\hline
\cellcolor[HTML]{C0C0C0}\textbf{2} & 0.43 & 0.92 & 0.08 \\
\hline
\cellcolor[HTML]{C0C0C0}\textbf{4} & 0.98 & 0.99 & 0.01 \\
\hline
\cellcolor[HTML]{C0C0C0}\textbf{5} & 0.83 & 0.99 & 0.01 \\
\hline
\cellcolor[HTML]{C0C0C0}\textbf{6} & 0.97 & 1 & 0 \\
\hline
\cellcolor[HTML]{C0C0C0}\textbf{7} & 0.69 & 0.99 & 0.01 \\
\hline
\cellcolor[HTML]{C0C0C0}\textbf{10} & 0.93 & 0 & 1 \\
\hline
\cellcolor[HTML]{C0C0C0}\textbf{11} & 0.72 & 0.97 & 0.03 \\
\hline
\cellcolor[HTML]{C0C0C0}\textbf{12} & 0.85 & 0.90 & 0.10\\
\hline
\cellcolor[HTML]{C0C0C0}\textbf{15} & 0.73 & 0.99 & 0.01\\
\hline
\cellcolor[HTML]{C0C0C0}\textbf{17} & 0.78 & 0.88 & 0.12\\
\hline
\cellcolor[HTML]{C0C0C0}\textbf{18} & 0.83 & 0.97 & 0.03\\
\hline
\cellcolor[HTML]{C0C0C0}\textbf{22} & 0.94 & 0.98 & 0.02\\
\hline
\cellcolor[HTML]{C0C0C0}\textbf{25} & 0.65 & 0.77 & 0.23\\
\hline

\end{tabular}
\label{table:evaluation_results_tests}
}
\end{table}

In this section we include an evaluation of the quality tests over the images of the second database, with the decision thresholds adjusted to the values obtained during the first evaluation. The main question we wanted to answer with this new analysis is: How consistent and discriminant are the results of the quality tests when executed on other type of face images?

Table~\ref{table:evaluation_results_tests} shows the performance of each individual test, including its accuracy, TPR, and FPR. We only calculated the results for the quality tests for which we were able to fix the decision threshold during the development evaluation. However, as mentioned previously, the database used in the second evaluation also presents a large mismatch between positive and negative classes for many tests, so the results for those will not be fully reliable.

We can now look more in detail to the results of the three different categories of tests defined in development, i.e., \textit{High performance}, \textit{Medium performance}, and \textit{Low performance}. First, for the tests of the \textit{High performance} category, i.e., $4$, $5$, $6$, $12$, and $22$, the accuracy rates we obtained are again among the highest ones, all between $0.83$ and $0.98$, so it seems acceptable to assume that these tests are generalizing correctly over new images. However, for the tests number $4$, $5$, and $6$, Table~\ref{table:test_db_labels} shows that that there is a practical absence of negative cases in this second database so their accuracy values must be taken with caution. Additionally, if we focus our analysis of those tests only on the images belonging to the positive class (from which we have a sufficient number of images to obtain statistically robust results) we can see that they are performing accurately with TPR values close to $0.99$ and FPR values close to $0.01$ in the three cases.

Secondly, if we analyse the results of the tests of the \textit{Low performance} category, it can be seen than the test number $2$ is the one with the worst accuracy, obtaining again a performance similar to random guess. Table~\ref{table:test_db_labels} showed us that this test presents a significant number of negative samples in the evaluation database, so this should not be affecting the process of computing its accuracy like it was in the case of other tests. For the other two quality tests in this category: $11$ (Pose) and $17$ (Light reflections on glasses), the accuracy is not too low (over $0.7$). In the case of the Pose test, the number of negative samples is really unbalanced with respect to the positive class so we think this can be causing that increase in accuracy respect to the development evaluation.

Summarizing, the results of the second evaluation show that the accuracy of the majority of the tests is consistent with the one obtained during development. Therefore, it seems acceptable to assume that the performance of the tests after the calibration of their decision thresholds is generalizable to other type of face images. The tests showed to be reliable to detect correctly those images that comply with specific points of the ISO/IEC standard, but to determine if they are equally accurate for detecting non-compliant images, an additional evaluation on a database with a high number of images with negative labels is necessary.

\section{Conclusion and future work}
\label{sec:conclusion}

In this paper we presented FaceQvec\footnote{https://github.com/uam-biometrics/FaceQvec}, a software tool that consists of $25$ tests related to face quality that check the conformance of face images with the requirements specified in the ISO/IEC 19794-5 standard. We included a brief description of each individual test, specifying the methods and the criteria we used for determining ISO/IEC conformance. 

We performed a first analysis on a development dataset where we calculated the performance of each one of the tests individually, and we used that information later to adjust their decision thresholds. After that, we evaluated FaceQvec again on a second database (applying the adjusted thresholds) and we verified that the face quality tests generalized correctly when facing new face images. 

Nevertheless, our analysis presents a limitation, i.e., the practical absence of negative samples for some quality tests both on the development and evaluation databases. This makes difficult to analyze the accuracy of the tests when facing non-compliant images. Additionally, there are some quality tests that are not as accurate as desirable, so further refinement work will help.

Derived from the previous conclusions, we propose as future lines of action to obtain a greater amount of balanced data for future analysis and better characterization of the system's performance. Taking advantage of new data, an improvement of the individual performance of some of the quality tests could be addressed in order to jointly contribute to an improvement in the overall system performance.

{\small
\bibliographystyle{ieee_fullname}
\bibliography{egbib}
}

\end{document}